\documentclass{article}
\usepackage{spconf,amsmath,graphicx}
\usepackage{amssymb}
\usepackage{bm}
\usepackage{xcolor}
\usepackage{dirtytalk}

\usepackage{caption}
\usepackage{subcaption}

\usepackage[normalem]{ulem}

\title{Automatic inspection of cultural monuments using deep and tensor-based learning on hyperspectral imagery}
\name{
\begin{tabular}{c@{}c@{}}
	Ioannis N. Tzortzis$^1$, 
	Ioannis Rallis$^1$, 
	Konstantinos Makantasis$^2$, 
	Anastasios Doulamis$^1$, \\ 
	Nikolaos Doulamis$^1$ and
	Athanasios Voulodimos$^1$
\end{tabular}\thanks{This work has been supported by the European Union's Horizon 2020 research and innovation programme from the TAMED project (Grant Agreement No. 101003397) and the H2020 YADES project, grant agreement No. 872931.}}

\address {1: National Technical University of Athens, Computer Vision and Photogrammetry Laboratory \\
2: University of Malta, Department of Artificial Intelligence}
\begin{document}
%\ninept
%
\maketitle
\begin{abstract}
In Cultural Heritage, hyperspectral images are commonly used since they provide extended information regarding the optical properties of materials. Thus, the processing of such high-dimensional data becomes challenging from the perspective of machine learning techniques to be applied. In this paper, we propose a Rank-$R$ tensor-based learning model to identify and classify material defects on Cultural Heritage monuments. In contrast to conventional deep learning approaches, the proposed high order tensor-based learning demonstrates greater accuracy and robustness against overfitting. Experimental results on real-world data from UNESCO protected areas indicate the superiority of the proposed scheme compared to conventional deep learning models.

\end{abstract}
%
% Rallis + Nikos

\section{Introduction}
\label{sec:intro}
%is the dataset hyperspectral or multispectral?
Cultural Heritage (CH) assets suffer from man-made hazards, and natural disasters \cite{Papachristou,Rallis} as UNESCO declares in \say{World Heritage in Danger} \cite {UNESCO}. Therefore, CH entities require regular inspection for defects, material deterioration and structure deformation. Currently, inspection is performed on-site and  manually by experts. However, this is a tedious, time consuming and costly task. Motivated by the advances in hyperspectral sensing and deep learning, in this paper, we propose \textit{an automatic non-invasive approach} for CH assets inspection.  

A typical deep learning model includes a large amount of tunable parameters to be learnt implying, in the sequel, a large amount of training samples. However, collection of labeled data is an expensive and tedious process, especially for CH applications. This is due the fact that (i) complementary approaches (invasive, non-invasive) should be carried out to verify the type and the degree of material deterioration and/or structural deformation and (ii) the captured hyperspectral image data should be annotated by engineers' experts. %Another issue of deep learning is that the models operate in a \say {black box} mode, meaning that there is no direct interpretation of how each spectral band (of hyperspectral imaging) contributes to the final classification.

Recently, \textit {tensor-based learning} has been emerged as a powerful alternative for classifying hyperspectral image data \cite{makantasis2018tensor,makantasis2021rank,makantasis2021space,makantasis2019common}. These works (i) apply a canonical decomposition of the model parameters to reduce the number of trainable weights---and thus the number of samples required for efficient training---and (ii) retain the raw (tensor) form of the data to fully exploit its structural information encoded across the available data modes (dimensions). Currently tensor learning have been mostly applied for remote sensing agricultural scenarios from satellite data. It should be mentioned that the scale of a remote sensing data is of order of Kms, making it impossible for detecting defects in CH assets. These works are motivated by the property of hyperspectral imaging to act as a \textit {material detector}  since different materials absorb or reflect light differently.
However, CH asset inspection requires high discrimination sensitivity since the model is designed to distinguish small deformations, defects and cracks which could be catastrophic for its stability.  Additionally, the scale of the analysis is of order of few centimeters imposing challenges both for performance and computational demands.  

\begin{figure*}
    \centering
    \includegraphics[scale=0.35]{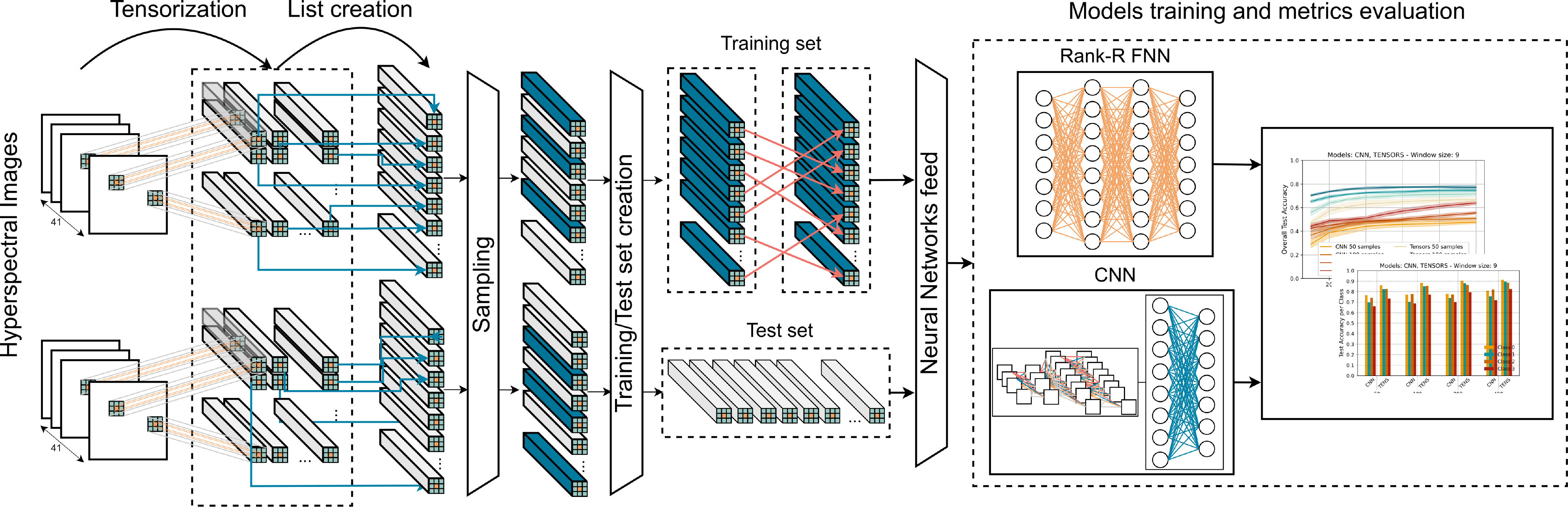}
    \caption{The proposed framework for automatic defect detection on CH assets using an Rank-$R$ tensor-learning model.}
    \label{fig:preproc}
\end{figure*}

In this study, we introduce a tensor-based learning system capable of detecting and classifying defects on CH monuments. Our proposed system exploits the Rank-$R$-FNN architecture \cite{makantasis2021rank} to overcome the main drawback of deep learning models, related to the requirements for large amounts of training data, without compromising detection accuracy. Our system detects and classifies defects using hyperspectral images captured with ground senors giving the opportunity to civil engineers to assess in a non invasive way and in real-time the impact of defects on the structure of CH monuments.

%For this reason, in this paper we propose  a new higher order (Rank-$R$) canonical decomposition of a tensor-based learning scheme instead of first order paradigm such as in \cite{makantasis2018tensor}. Rank-$R$ canonical decomposition increases discrimination capabilities of the learning models, for detecting contaminated materials and defects at few cm scale. This is in contrast to traditional agricultural applications where the target is to detect quite different material types of land properties at Km scale. In addition, our proposed scheme is designed only to detect the presence of a defect but also to classify the defect type. Material classification is a critical factor for civil engineers to assess the impact of a defect on structure of a monument.  We should stress that we utilise ground hyperspectral imaging sensors and multi-spectral cameras o UAVs instead of satellite remote sensing data. 

To sum up, our paper is novel at the following directions. First, it introduces a high-order tensor-based learning system using ground hyperspectral data for defect detection and classification in CH structures. Second, it utilizes much smaller training samples than other conventional deep learning approaches for training, significantly reducing annotation effort and computational demands. Third, it introduces a new real-world dataset captured at UNESCO protected areas for evaluating the proposed scheme.

\subsection{Related Works}
\label{sec:format}
%This section summarizes literature related to our study across two different directions; defect detection systems that exploit hyperspectral data, and current CH assets inspection techniques.

\textbf{Defect detection with hyperspectral data:}
The authors in \cite{alexopoulou} show that spectral imaging techniques can offer several possibilities in exploiting optical properties of materials, by taking into account specific characteristics of the spectral bands. The study in \cite{7274902} tries to address the problem of crack detection on paintings. Towards this direction, distance-based spectral mathematical morphology was used, offering a vector and full-band processing approach. Additionally, several top-hat transformations were utilized to assist the task. This work also confirms the capacity of hyperspectral imaging to offer additional useful and essential information. However, none of the aforementioned approaches is married with deep and tensor-based learning schemes to improve classification performance across difficult real-world CH paradigms and being able to discriminate not only the defects but also the defect types as we perform in presented study.  

\textbf{Inspection of CH assets:}
The study in \cite{9506300} describes a data fusion pipeline for the delayering of X-ray fluorence (XRF) images. At first, visible hyperspectral reflectance data (RIS) is clustered in pigment mixtures. Then, a synthetic surface XRF image is formed by calculating the mean XRF response across all clusters. Finally, the surface and subsurface correlated features are identified by subtracting the synthetic surface XRF from the full image. In \cite{polak2017hyperspectral}, the combination of hyperspectral imaging with advanced signal processing techniques is proposed as a tool to assist the artwork authentication procedure. According to the authors, this is achieved by applying classification techniques on coloured pigments. The Support Vector Machine (SVM) algorithm was used in combination with the ``one-against-one'' technique in order to facilitate the multi-class problem. In \cite{kolokoussis20213d}, the authors exploit 3D textured  models in combination of hyperspectral imagery to define  specific areas of degradation. Although these works utilize state-of-the-art deep learning techniques, they require a large amount of samples for the training process. In stark contrast, the proposed tensor-based learning model requires noticeably lower amount of training samples. 

\begin{figure}
    \centering
    \includegraphics[width=0.32\linewidth]{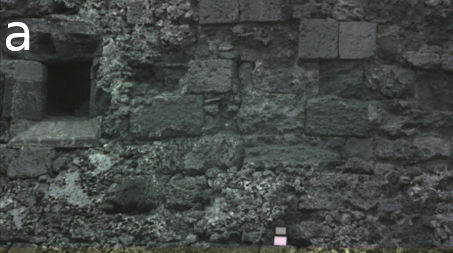}
    \includegraphics[width=0.32\linewidth]{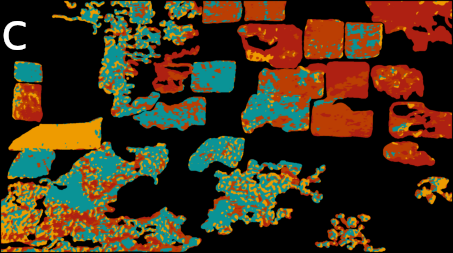}
    \includegraphics[width=0.32\linewidth]{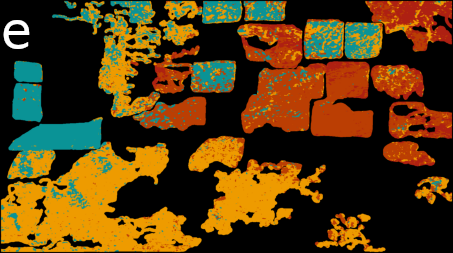}
    \includegraphics[width=0.32\linewidth]{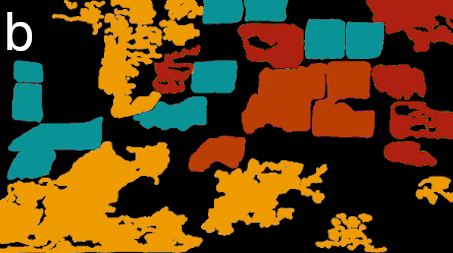}
    \includegraphics[width=0.32\linewidth]{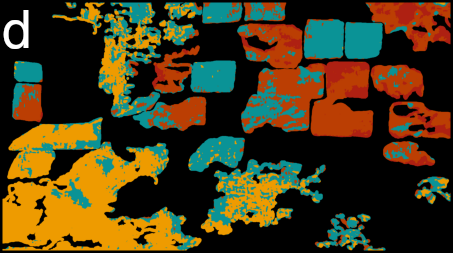}
    \includegraphics[width=0.32\linewidth]{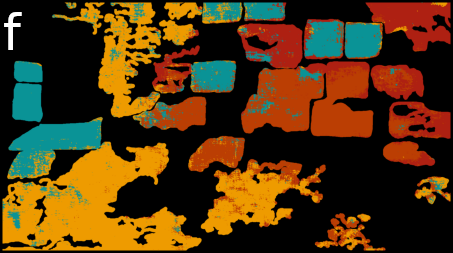}
    \includegraphics[width=1\linewidth]{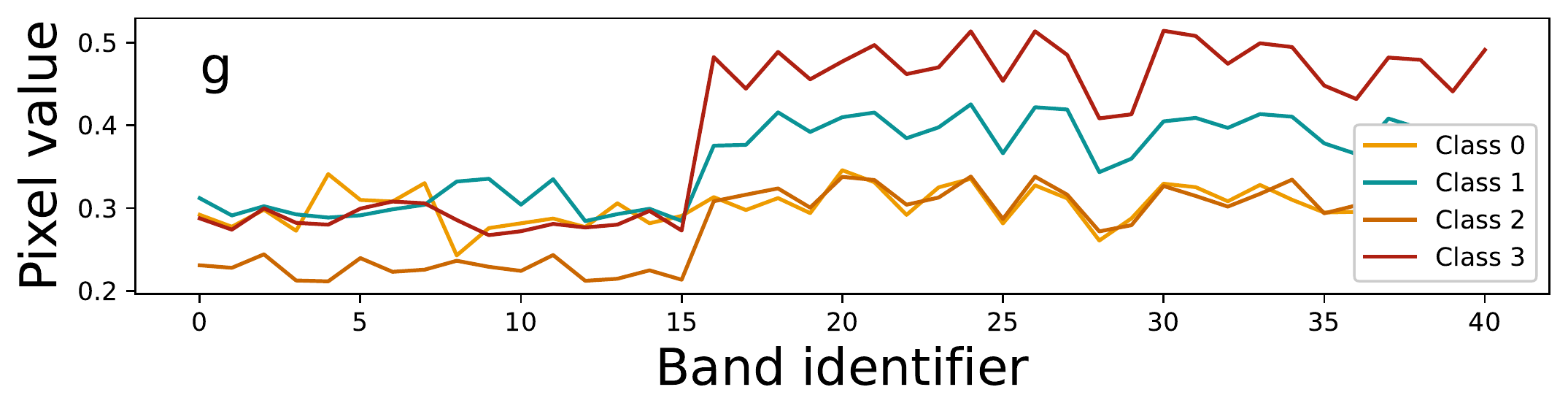}
    \caption{a) Sample image from the dataset, b) the corresponding annotation, c) CNN prediction (TWS=9, TS=50), d) CNN prediciton (TWS=21, TS=400), e) Rank-$R$ FNN prediciton (TWS=9, TS=50), f) Rank-$R$ FNN prediciton (TWS=21, TS=400), g) Spectral response of representative pixels from each class.}
    \label{fig:dataset}
\end{figure}

% \begin{figure*} [ht!]
%     \centering
%     \includegraphics[trim={0.4cm 0cm 1.5cm 0.4cm}, clip, width=0.32\textwidth]{results/CNN_TENS_9.eps}
%     \label{subfig:ov}
%     \includegraphics[trim={0.4cm 0cm 1.5cm 0.4cm}, clip, width=0.32\textwidth]{results/CNN_TENS_15.eps}
%     \includegraphics[trim={0.4cm 0cm 1.5cm 0.4cm}, clip, width=0.32\textwidth]{results/CNN_TENS_21.eps}
%     \includegraphics[trim={0.4cm 0cm 1.5cm 0.4cm}, clip,width=0.32\textwidth]{results/CNN_TENS_9class.eps}
%     \includegraphics[trim={0.4cm 0cm 1.5cm 0.4cm}, clip,width=0.32\textwidth]{results/CNN_TENS_15class.eps}
%     \includegraphics[trim={0.4cm 0cm 1.5cm 0.4cm}, clip,width=0.32\textwidth]{results/CNN_TENS_21class.eps}
%     \caption{In the top row, the graphs show the 95\% confidential interval of the overall test accuracy for both models and different adjustments of TS and TWS parameters. In the bottom row, the mean accuracy per class is presented, for both models and different adjustments of TS and TWS parameters.}
%     \label{fig:overall_acc}
% \end{figure*}

\begin{figure*}
     \centering
     \captionsetup[subfigure]{aboveskip=-0.6pt,belowskip=-1pt}
     \begin{subfigure}[b]{0.3\textwidth}
         \centering
         \includegraphics[trim={0.4cm 0cm 1.5cm 0.4cm}, clip, width=1\textwidth]{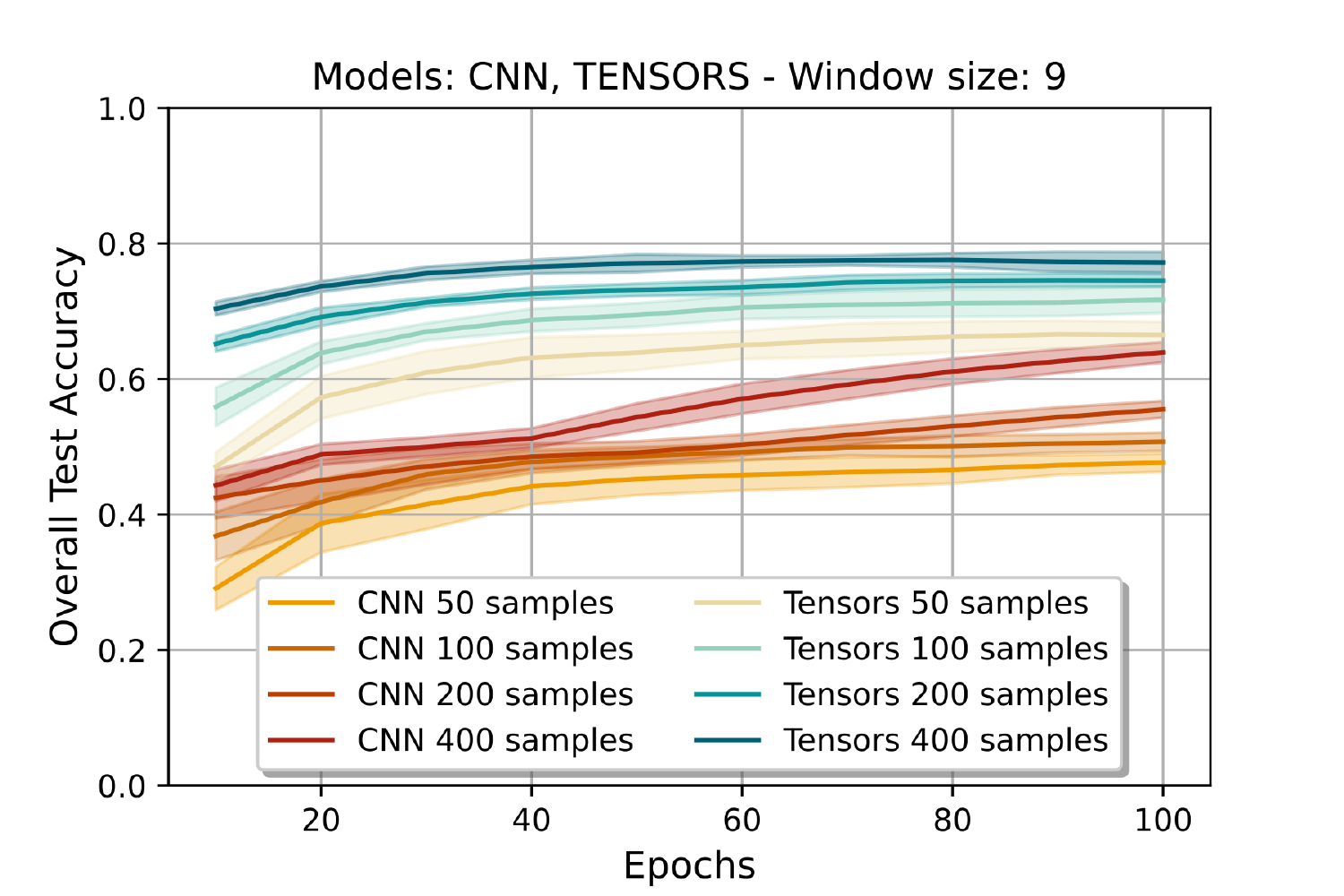}
         \caption{}
         \label{fig:ct9}
     \end{subfigure}
     %\hfill
     \begin{subfigure}[b]{0.3\textwidth}
         \centering
         \includegraphics[trim={0.4cm 0cm 1.5cm 0.4cm}, clip, width=1\textwidth]{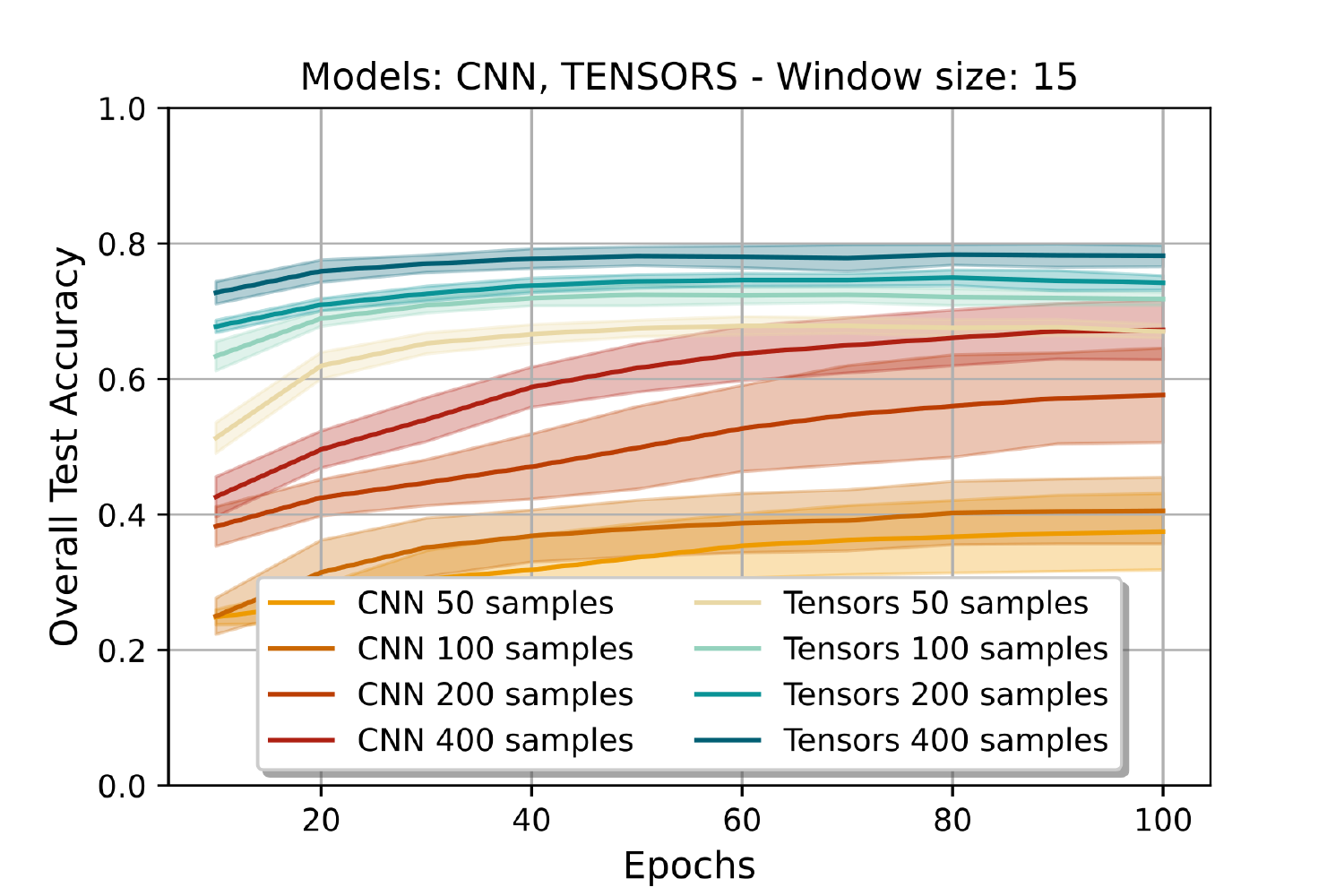}
         \caption{}
         \label{fig:ct15}
     \end{subfigure}
     %\hfill
     \begin{subfigure}[b]{0.3\textwidth}
         \centering
         \includegraphics[trim={0.4cm 0cm 1.5cm 0.4cm}, clip,width=1\textwidth]{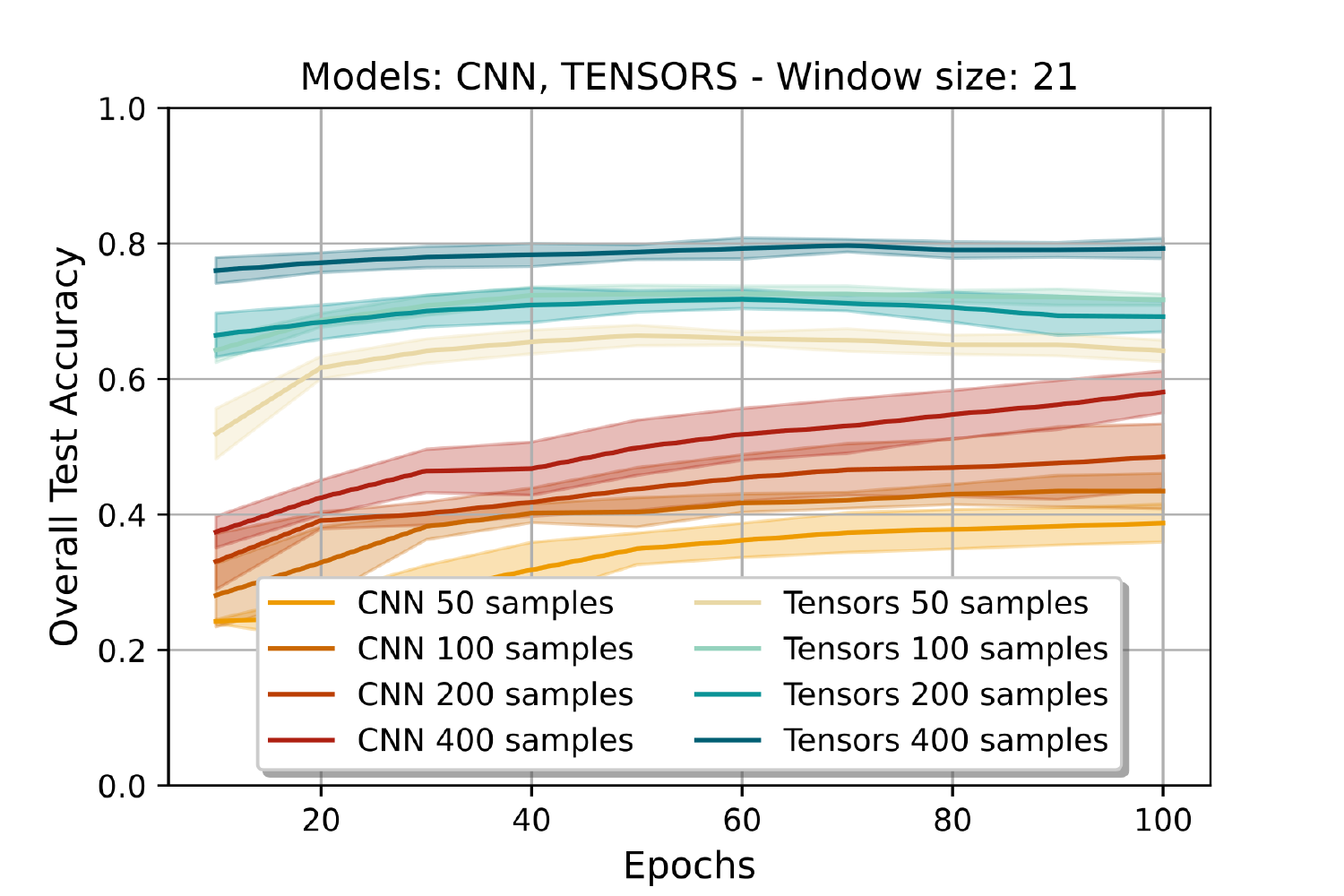}
         \caption{}
         \label{fig:ct21}
     \end{subfigure}
     \begin{subfigure}[b]{0.3\textwidth}
         \centering
         \includegraphics[trim={0.4cm 0cm 1.5cm 0.4cm}, clip, width=1\textwidth]{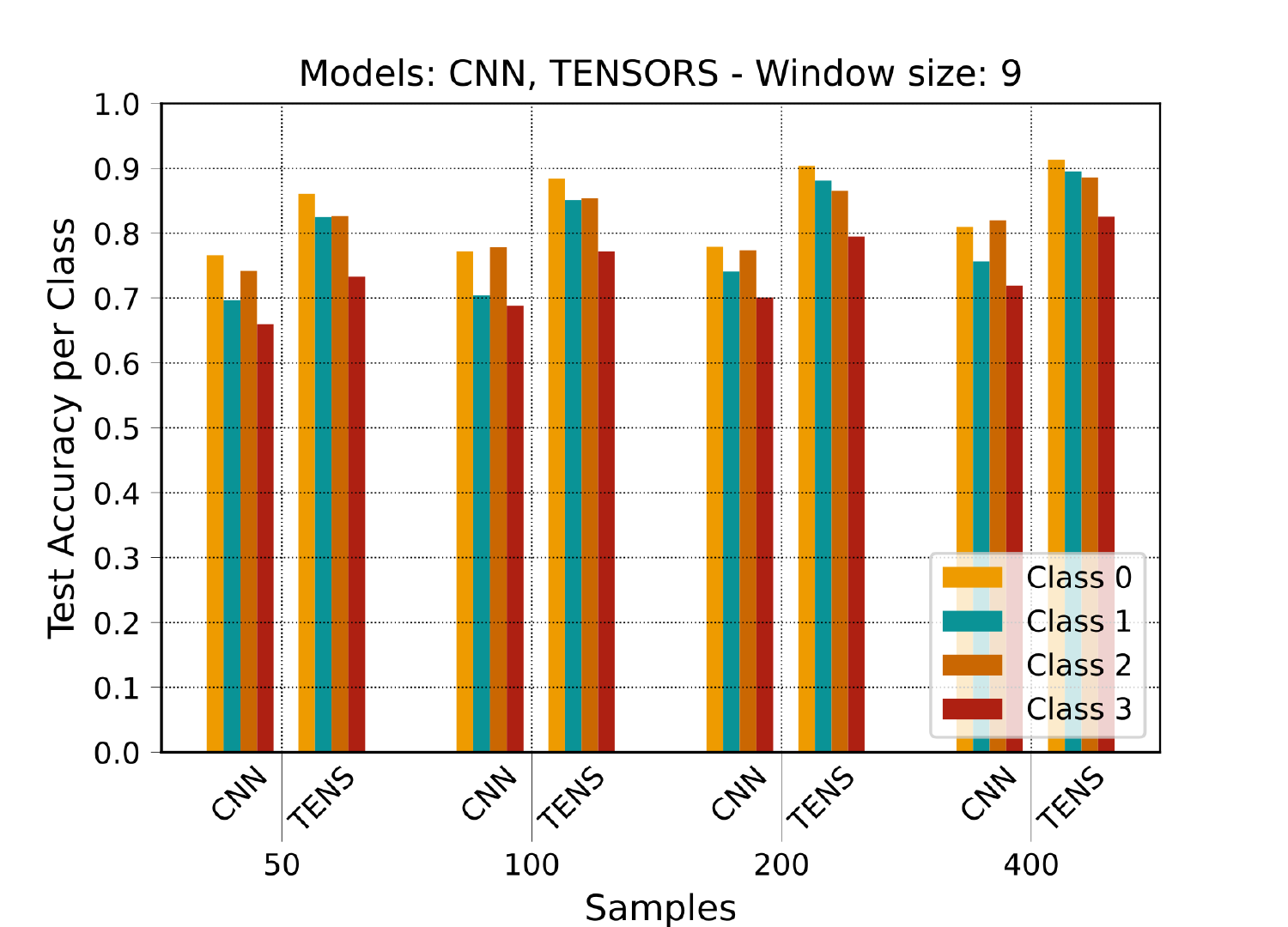}
         \caption{}
         \label{fig:ct9c}
     \end{subfigure}
     %\hfill
     \begin{subfigure}[b]{0.3\textwidth}
         \centering
         \includegraphics[trim={0.4cm 0cm 1.5cm 0.4cm}, clip, width=1\textwidth]{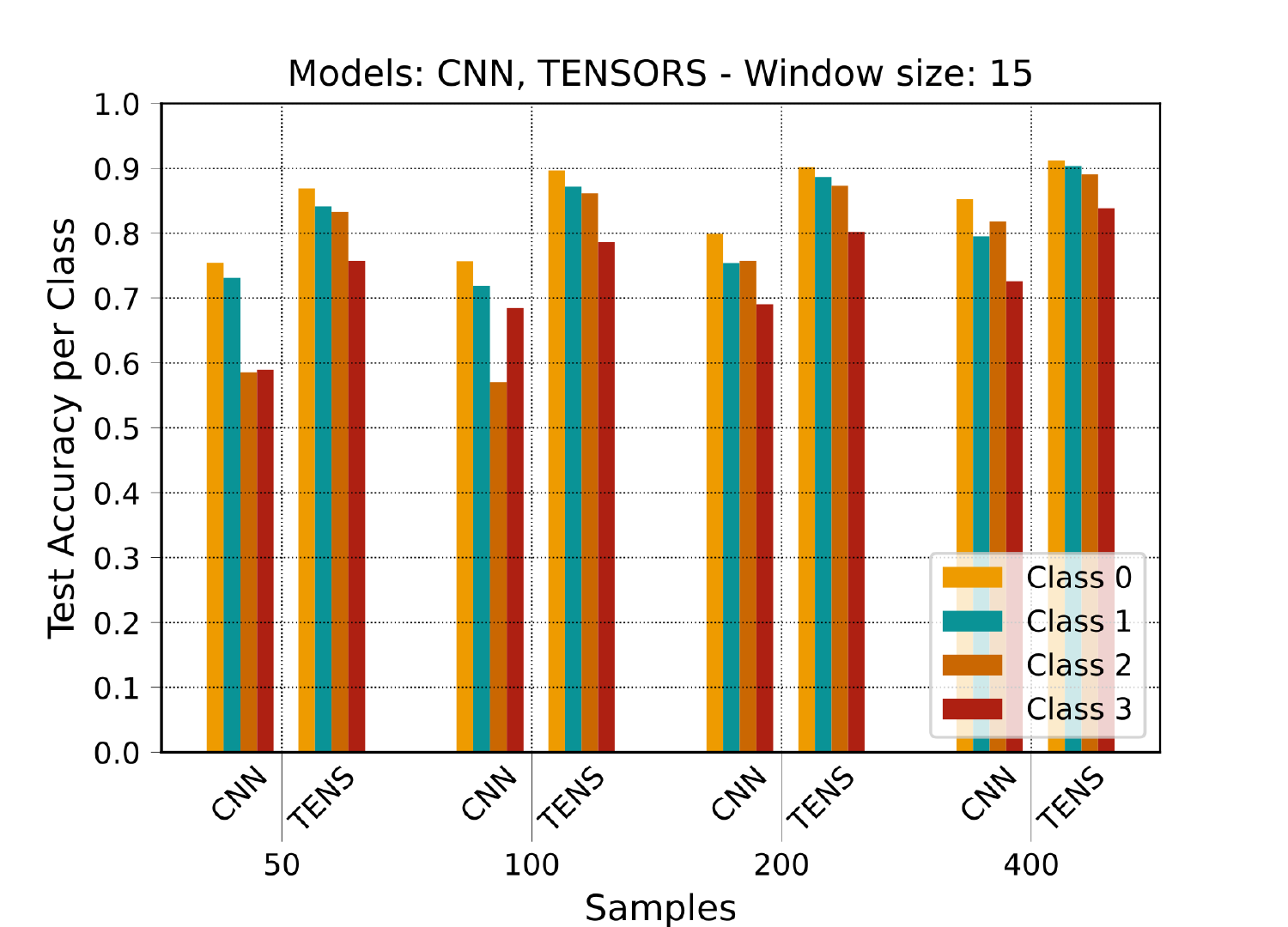}
         \caption{}
         \label{fig:ct15c}
     \end{subfigure}
     %\hfill
     \begin{subfigure}[b]{0.3\textwidth}
         \centering
         \includegraphics[trim={0.4cm 0cm 1.5cm 0.4cm}, clip,width=1\textwidth]{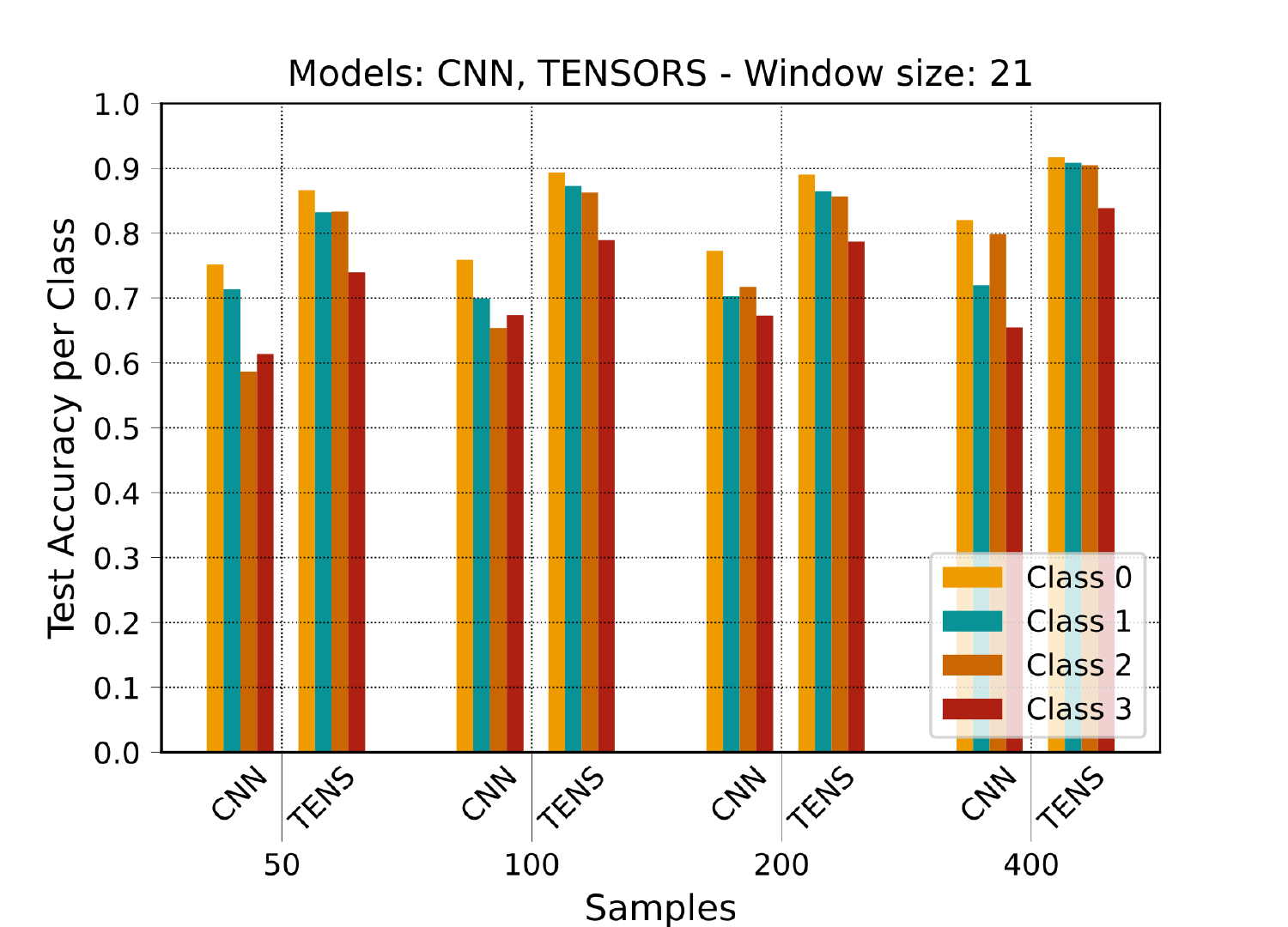}
         \caption{}
         \label{fig:ct21c}
     \end{subfigure}
     
        \caption{In the top row, the graphs show the 95\% confidential interval of the overall test accuracy for both models and different adjustments of TS and TWS parameters. In the bottom row, the mean accuracy per class is presented, for both models and different adjustments of TS and TWS parameters.}
        \label{fig:exps}
\end{figure*}

\section{Methodology}
\label{ProblemFormulation}
%In this section we, first, formulate the problem of defect detection using hyperspectral imagery and, then, we present the approach we follow to efficiently address it. 

\subsection{Problem Formulation}
Our problem can be seen as a \textit {multi-class classification} task. Let us denote as $C$ the number of classes. Then, we can detect $C-1$ different types of defects plus one class of pixels that correspond to no defect.  

Let us denote as $X_i \in \mathcal X$ the information describing the $i$-th pixel of a hyperspectral image, and as $t_i = [t_{i,1}, ...,t_{i,k}, ..., t_{i,C}]^T \in \{0,1\}^C$, such that $\sum_{j=1}^C t_{i, j} = 1$, the ground truth label vector for the same pixel. Then given a collection
\begin{equation}
\label{eq:dataset}
    \mathcal D = \{(X_i, t_i)\}_{i=1}^N
\end{equation}
the problem of automatic defect detection boils down to estimate a function $f(\cdot, \theta^*):\mathcal X \rightarrow \{0,1\}^C$, such that
\begin{equation}
\label{eq:loss}
    \theta^* = \arg \min_{\theta \in \Theta} \frac{1}{N} \sum_{i=1}^N L_{CE} \big(f(X_i, \theta), t_i\big). 
\end{equation}
In Eq. (\ref{eq:loss}), $L_{CE}$ stands for the cross-entropy loss and $\theta \in \Theta$ is the set of parameters that determine the form of $f$.

\subsection{Rank-$R$ Tensor Based Learning}
\label{ssec:approach}
To efficiently detect defects, each $X_i$ should carry both spectral and spatial information describing the $i$-th pixel. In other words, each $X_i$ should carry information about the spectral response of pixel $i$ and information about $i$-th pixel's neighbors. Towards this direction, with $X_i$ we represent a square patch of dimensions $s \times s \times b$ of the hyperspectral image centered at the $i$-th pixel. Parameter $s$ stands for the height and width of the patch, and $b$ for the number of spectral bands. This way, each $X_i \in \mathbb R^{s \times s \times b}$ is a 3rd-order tensor encoding both the spatial and spectral information of pixel $i$.

To address the problem formulated in the previous section, we represent the function $f(\cdot, \theta)$ by a tensor-based machine learning model utilizing high-order canonical decomposition. We call this model as Rank-$R$ Feed Forward Tensor-based Neural Network (Rank-$R$ FNN) since it exploits tensor operators in a feedforward structure. Particularly,   Rank-$R$ FNN is a neural network with one hidden layer which consists of, let's say, $Q$ hidden neurons. Rank-$R$ FNN weights connecting the input to hidden layer are tensors satisfying the Rank-$R$ Canonical-Polyadic decomposition \cite{kolda2009tensor}:

%Specifically, we utilize a network model along with a Rank-$R$  

%use two different types of neural networks; a convolutional neural network model, called Tensor-CNN \cite{makantasis2015deep} and a Rank-$R$ Feed Forward Tensor-based Neural Network \cite{makantasis2018tensor, makantasis2021rank}, called Rank$R$ FNN. Both network structures have been used only for land cover classification using hyperspectral data. Tensor-CNN applies a sequence of convolution operations combined with a multilinear perceptron and it produces high-level representations of hyperspectral data and achieves very accurate classification results.  

\begin{equation}
    \bm w^{(q)} = \sum_{k=1}^R \bm w_{3,k}^{(q)} \circ \bm w_{2,k}^{(q)} \circ \bm w_{1,k}^{(q)} \in \mathbb R^{b \times s \times s}, 
\end{equation}
for $q=1,\cdots,Q$ with $\bm w_{3,k}^{(q)} \in \mathbb R^b$ and $\bm w_{i,k}^{(q)} \in \mathbb R^s$, $i=1,2$. Superscript $q$ denotes that these weights connect the input to the $q$-th neuron of the hidden layer, and ``$\circ$'' operator stands for vectors outer product. The output of the Rank-$R$ FNN for the $c$-th class is 
\begin{equation}
    p^c = \sigma(\langle \bm v^{(c)}, \bm u \rangle),
\end{equation}
where $\bm v^{(c)}$ collects the weights between the hidden layer and the $c$-th output neuron, $\sigma(\cdot)$ denotes the softmax activation function, and $\bm u=[u_1, u_2, \cdots, u_Q]^T$ with
\begin{equation}
    u_q = g\bigg(\bigg\langle \Big(\bm \sum_{k=1}^R \bm w_{3,k}^{(q)} \circ \bm w_{2,k}^{(q)} \circ \bm w_{1,k}^{(q)} \Big), X_i\bigg\rangle\bigg) 
\end{equation}
for $q=1,\cdots,Q$ to be the output of the hidden layer activated by function $g(\cdot)$. In this study, we use and compare the Rank-$R$ FNN and the CNN in \cite{makantasis2015deep} since both models exploit spatio-spectral pixels' information and can be used for pixel-wise hyperspectral image classification tasks. Given a collection of training data in the form of relation (\ref{eq:dataset}), we estimate the set of parameters of the employed models using the backpropagation algorithm \cite{lecun2015deep} with the Adam gradient based optimizer \cite{kingma2014adam}. Fig.\ref{fig:preproc} presents our overall approach.

%% Tzortzis
\section{Experimental Results}
\label{sec:experiments} 

\subsection{Dataset description}
We use a new dataset consisting of hyperspectral images depicting part of ancient walls of Saint Nicolas fortress located in the UNESCO Heritage Medieval city of Rhodes, Greece. The 6 images of the dataset were collected using the HyperView sensing platform \cite{garea2016hyperview} by 3D-one, which combines the information from one Visual (VIS) snap-shot camera and one Near Infrared (NIR) snap-shot camera. Each hyperspectral image consists of 1016 x 1820 pixels and 42 spectral bands. Moreover, each image is accompanied by a pixel-based annotated ground truth image (see Fig. \ref{fig:dataset}.b), carried out by CH experts. Four different classes are considered as is depicted in Fig. \ref{fig:dataset} with different colors; \textit{class 0} represents the salt defects (yellow), \textit{class 1} shows the non-significantly defected areas(light blue), \textit{class 2} depicts minor deterioration (orange) and \textit{class 3} (red) shows major deterioration. Fig. \ref{fig:dataset}.g presents the spectral response of representative pixels from each class.
%@Tzortzis: is there a no-defect class?

%The dataset, used for the purposes of this paper, consists of 6 hyper-spectral images that depict part of ancient walls, part of the greek cultural heritage.  Each image consists of 42 channels that correspond to specific frequencies of the electromagnetic spectrum. A group of ground truth images are attached to the original images of the dataset. Four different types of deterioration (classes) are included in the information provided by the ground truth images. In Figure \ref{fig:dataset}, a sample image is presented along with its corresponding ground truth labels. The white-coloured areas refer to class 0, the red-coloured areas to class 1, the blue-coloured areas to class 2 and the green-coloured areas to class 3.

\subsection{Pre-processing pipeline}
The hyperspectral images are partitioned into tensor objects. The sampling unit selects random samples from the tensor objects  to create the training and the test sets. A permutation process is also applied on the training set to reduce bias effect. We normalize each hyperspectral image in a band-wise manner using the min-max normalization to restrict the pixels' responses at each band to $[0,1]$. After normalization, we split each image into patches of dimension $s \times s \times 42$ (tensorization step in Fig.\ref{fig:preproc}). During our experiments we set parameter $s$ equal to 9, 15 and 21 to investigate the effect of patch height and width on the defect detection accuracy. The generated patches of all the available hyperspectral images are aggregated into a single set. From this set, we randomly select a number samples per class for training the models (the proposed tensor based and compared), while the rest of the patches are used as a test set for evaluating the defect detection performance. We have created the training set by randomly selecting 50, 100, 200 and 400 samples per class to to investigate the impact of training set size on learning models' performance. We repeat the aforementioned hold-out cross validation scheme 10 times and report the average model classification accuracy and 95\% confidence intervals.

\subsection{Performance Evaluation }
In this section, we evaluate the defect detection performance in terms of model classification accuracy and investigate the impact of patch size (parameter $s \in \{9, 15, 21\}$) and training set size (50, 100, 200 and 400 samples per class). Two essential hyper-parameters are examined; \textit{Tensor Window Size (TWS)} and \textit {Tensor Samples (TS)}. Twelve well defined different experiments are designed and performed. Three distinct values of TWS (9, 15 and 21) are used and for each value, four increasing values of TS are selected (50, 100, 200 and 400). The models are trained for 100 epochs and the results are calculated on the test set. Each experiment is repeated ten times to ensure the validity of the results. 

In Fig. \ref{fig:ct9}-\ref{fig:ct21} we present the overall accuracy of the proposed tensor based model compared with the state-of-the-art CNN-based model used in the literature for hyperspectral image classification \cite{makantasis2015deep}.The accuracy is presented for different number of training samples (TS range from 50 to 400 samples / class) and TWS ranging from 9 to 21. In this figure, the colored shaded areas across the lines stand for the standard deviation value. As is observed, the tensor based model presents the best accuracy, while retaining its performance  for small number of training samples. In particular, for TWS=9, the accuracy of the state-of-the-art CNN varies from  45\% and 65\% depending on the TS values, while the accuracy of the proposed tensor model ranges from 65\% to 78\%. It should be mentioned that as the number of training samples (TS) increases the accuracy of the CNN is similar to the tensor model. However, the CNN model presents much higher standard deviation across the several repetitions of the experiments, implying a much lower robustness compared to the proposed tensor model. Fig. \ref{fig:ct9c}-\ref{fig:ct21c} presents the accuracy of the proposed tensor model and the compared CNN across the different defect classes and different TWS values 9, 15 and 21 respectively. As is observed, the proposed tensor model outperforms the CNN approach over the different classes. 

In Fig. \ref{fig:dataset}, a visual representation of both models predictions is demonstrated, along with the original image and the corresponding ground truth annotation. The results are depicted fro different number of training samples (TS) and window sizes (TWS). Even the worst case of Rank-$R$ FNN (Fig. \ref{fig:dataset}.e), which uses a small number of training samples, is more accurate than the best CNN approach (Fig. \ref{fig:dataset}.d). This reveals the robustness of the proposed tensor scheme for a small number of training samples.

\section{Conclusion}
\label{sec:illust}
In this work, we introduced the Rank-$R$ tensor-based learning model for the detection of different defect types on CH monuments using hyperspectral images. This model is compared against a state-of-the-art CNN. According to the results, the proposed method achieves higher accuracy score, even for low amount of training samples, while its standard deviation is lower than the CNN. In general, the proposed Rank-$R$ FNN increases the accuracy score 20\% more than the CNN approach.
\bibliographystyle{IEEEbib}
\bibliography{refs}

\end{document}